\documentclass{article}

\usepackage{arxiv}

\usepackage[utf8]{inputenc} 
\usepackage[T1]{fontenc}    
\usepackage[colorlinks=true]{hyperref}
\hypersetup{ 
    citecolor=black,%
    filecolor=black,%
    linkcolor=black,%
    urlcolor=black 
}
\usepackage{url}            
\usepackage{booktabs}       
\usepackage{amsfonts}       
\usepackage{nicefrac}       
\usepackage{microtype}      
\usepackage{cleveref}       
\usepackage{lipsum}         
\usepackage{graphicx}
\usepackage{natbib}
\usepackage{doi}
\usepackage[section]{placeins}

\title{GAIA: A General AI Assistant for Intelligent Accelerator Operations}

\date{}

\author{ \href{https://orcid.org/0000-0003-3816-0686}{\includegraphics[scale=0.06]{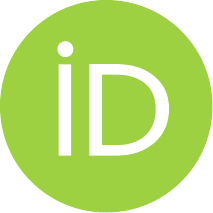}\hspace{1mm}Frank~Mayet}\\
	Accelerator R\&D (MPY1 Group)\\
	Deutsches Elektronen-Synchrotron DESY\\
	Notkestr. 85, 22607 Hamburg, Germany \\
	\texttt{frank.mayet@desy.de} \\
}

\renewcommand{\headeright}{Technical Report}
\renewcommand{\undertitle}{Technical Report}
\renewcommand{\shorttitle}{GAIA}

\hypersetup{
pdftitle={GAIA: A General AI Assistant for Intelligent Accelerator Operations},
pdfsubject={cs.CL, physics.acc-ph},
pdfauthor={Frank~Mayet},
pdfkeywords={Particle Accelerator Operations, LLM, Prompt Engineering},
}

\begin{document}
\maketitle

\begin{abstract}
	Large-scale machines like particle accelerators are usually run by a team of experienced operators. In case of a particle accelerator, these operators possess suitable background knowledge on both accelerator physics and the technology comprising the machine. Due to the complexity of the machine, particular subsystems of the machine are taken care of by experts, who the operators can turn to. In this work the reasoning and action (ReAct) prompting paradigm is used to couple an open-weights large language model (LLM) with a high-level machine control system framework and other tools, e.g. the electronic logbook or machine design documentation. By doing so, a multi-expert retrieval augmented generation (RAG) system is implemented, which assists operators in knowledge retrieval tasks, interacts with the machine directly if needed, or writes high level control system scripts. This consolidation of expert knowledge and machine interaction can simplify and speed up machine operation tasks for both new and experienced human operators.
\end{abstract}


\section{Introduction}
Particle accelerators are complex machines that consist of a large number of subsystems. Although many processes are automated and feedback systems are in place, experiments and machine supervision need to be performed by a group of operators. These operators usually have an accelerator physics background and mostly know how the technology works. They especially know how to setup and tune the machine parameters for certain working points and experiments using high-level graphical user interfaces, which are connected to low-level machine control software. Due to the complexity of the machine, some subsystems of the machine are taken care of by experts, who the operators can turn to. This work shows that it is possible to support the day-to-day operation of a complex machine like a particle accelerator using a large language model (LLM), an object-oriented high-level machine control system framework, as well as a number of interfaces to knowledge bases such as the electronic logbook. The system is able to assist the operators on many levels, e.g. by producing Python scripts, which when executed perform a task defined by an input prompt to the LLM. To this end, the reasoning and action prompting paradigm (ReAct) \citep{yao_react_2023} is implemented. This way a multi-expert system is realized, mimicking the real world, where the complex machine is operated by many subsystem experts.

\section{Background – Accelerator Controls}
Modern accelerators are routinely operated using sophisticated low-level control systems, such as EPICS\footnote{\url{https://epics.anl.gov}, last access 2024-04-26}, TANGO\footnote{\url{https://www.tango-controls.org}, last access 2024-04-26}, TINE\footnote{\url{https://tine.desy.de}, last access 2024-04-26}, or DOOCS\footnote{\url{https://doocs.desy.de}, last access 2024-04-26}. These control systems allow for high frequency machine and beam diagnostics, as well as control, which is an essential requirement for highly available, brilliant beams for the users. In addition, control systems are often paired with high speed data acquisitions systems (DAQ), capable of recording pulse synchronized data at the machine repetition rate. In order to access the machine, or derived data (via middle layer servers), users can usually leverage libraries and wrappers for common programming and scripting languages such as C, C++, Java, or Python. Easy to use toolkits are sometimes provided to create graphical user interfaces (GUIs) for simple interaction with control system servers (e.g. parameter scans), or data visualization.

On many machines there is the need to perform experiments, which are more complex than simple one- or two-dimensional parameter scans. In this case users and operators have to either perform the tasks manually, or to write measurement scripts using e.g. Python. In this work a Python-based scripting toolkit called \texttt{doocs\_generic\_experiment} is used, which simplifies the task of writing a measurement script substantially. The toolkit follows an object-oriented approach and is based on a set of so called \emph{actions}, which resemble real-world sub-tasks an operator would have to perform if the experiment would be conducted manually. These actions can be grouped into \emph{procedures}, which can reach arbitrary complexity. In addition of performing a set of action in series, procedures can also run multiple actions in parallel to save time. The final experiment can then consit of either one, or many of these procedures. 

\begin{figure}[ht]
	\centering
	\includegraphics[width=0.6\columnwidth]{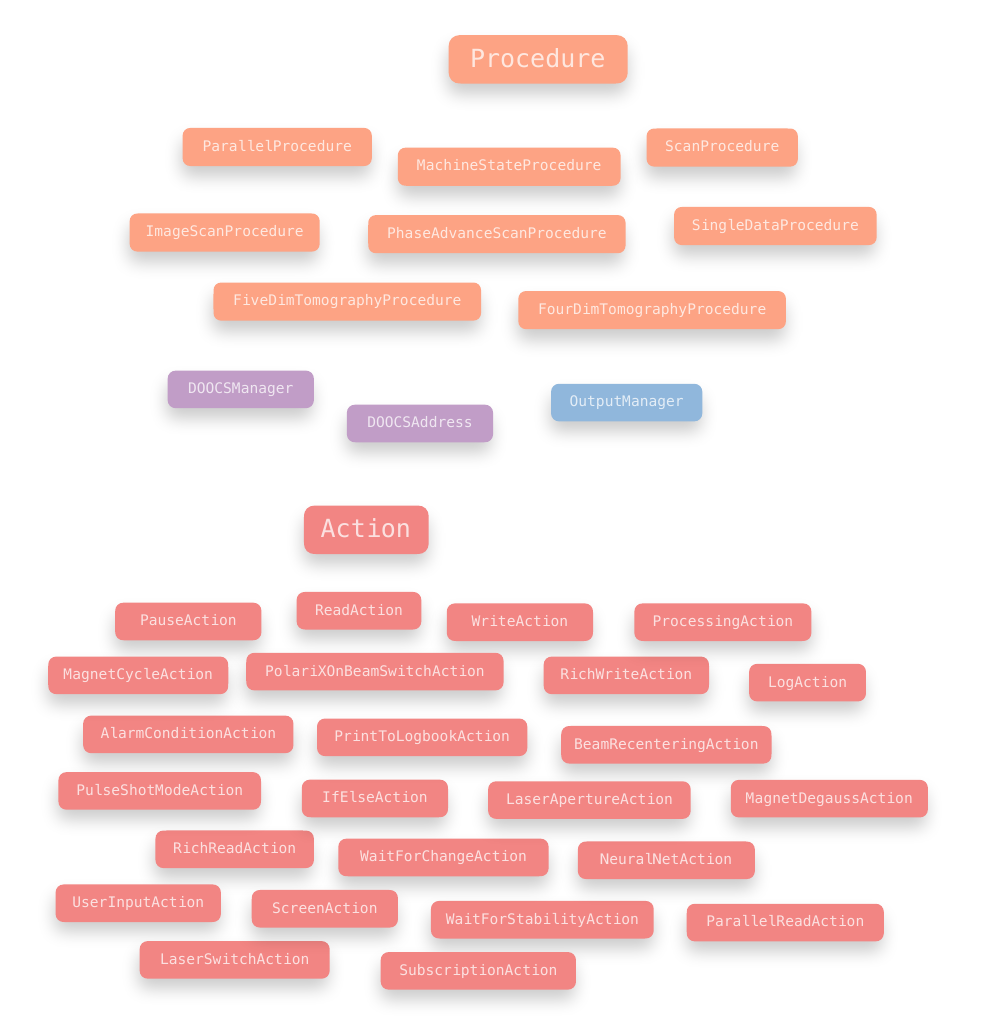}
	\caption{A selection of procedures and actions available via the \texttt{doocs\_generic\_experiment} Python module.}
	\label{fig:dge_selection}
\end{figure}

Due to the easy to unterstand concept of stringing together common actions and procedures, the toolkit enables rapid prototyping of complex experiments and enables full- and semi-automation of experimental campaigns, which would otherwise take too long to perform manually. In addition, the concept of encapsulating commonly performed actions adds a safety layer as the room for possible mistakes is reduced. Figure~\ref{fig:dge_selection} shows a selection of currently available actions and procedures.

\section{Multi-Expert System – GAIA}
In order to realize the operations assistant, the open-weights LLM Mixtral 8x7B Instruct v0.1 (8 bit quantization) \citep{jiang_mixtral_2024} is used as the main model. Mixtral 8x7B supports a context size of 32k tokens. This is ideal for reasoning and chain of thought (CoT) prompting \citep{wei_emergent_2022,wei_chain--thought_2023,yao_react_2023}. The model runs locally on a single Nvidia A100 80GB GPU in the DESY Maxwell cluster and is served by \texttt{Ollama}\footnote{\url{http://www.ollama.com}, last access 2024-04-26} (model tag: \texttt{mixtral:8x7b-instruct-v0.1-q8\_0}). Note that the Maxwell node does not need to directly interface with the accelerator control system. All interaction and knowledge retrieval is performed by a client application, which runs on a computer, which is part of the control system network. This client can then interface with control system servers, file servers, the Mattermost messaging system, etc., if needed. The client is called \emph{General AI Assistant} (GAIA).

One of the main concepts of the ReAct prompting scheme is tool use. As the agent engages in its inner monologue, or chain of thought, it will eventually reach a point where it needs to either perform an action, or retrieve information. This is analogous to an operator deciding to either use control system tools to control or diagnose certain parts of the machine, or to turn to an expert to retrieve information. Technically, the agent, which is implemented using LangChain\footnote{\url{http://www.langchain.com}, last access 2024-04-26}, interrupts the LLM output stream and injects the result of a particular tool use. The agent might, for example, come to the conclusion that it needs to retrieve the current value of a particular machine parameter. In this case, within its chain of thought, it may suggest using \texttt{doocs\_generic\_experiment} based tools to perform this particular task. Another example would be the retrieval of information from the electronic logbook. 

One advantage of the concept of tool use is that it helps to circumvent the LLM token limit. In many cases knowledge retrieval may involve classical retrieval augmented generation (RAG) \citep{lewis_retrieval-augmented_2021} implementations, which use their own disjunct LLM context. This way, only the result of the knowledge retrieval process is injected into the agents context window. In addition, this allows the use of fine-tuned RAG systems potentially based on other LLMs.

\begin{figure}[ht]
	\centering
	\includegraphics[width=0.9\columnwidth]{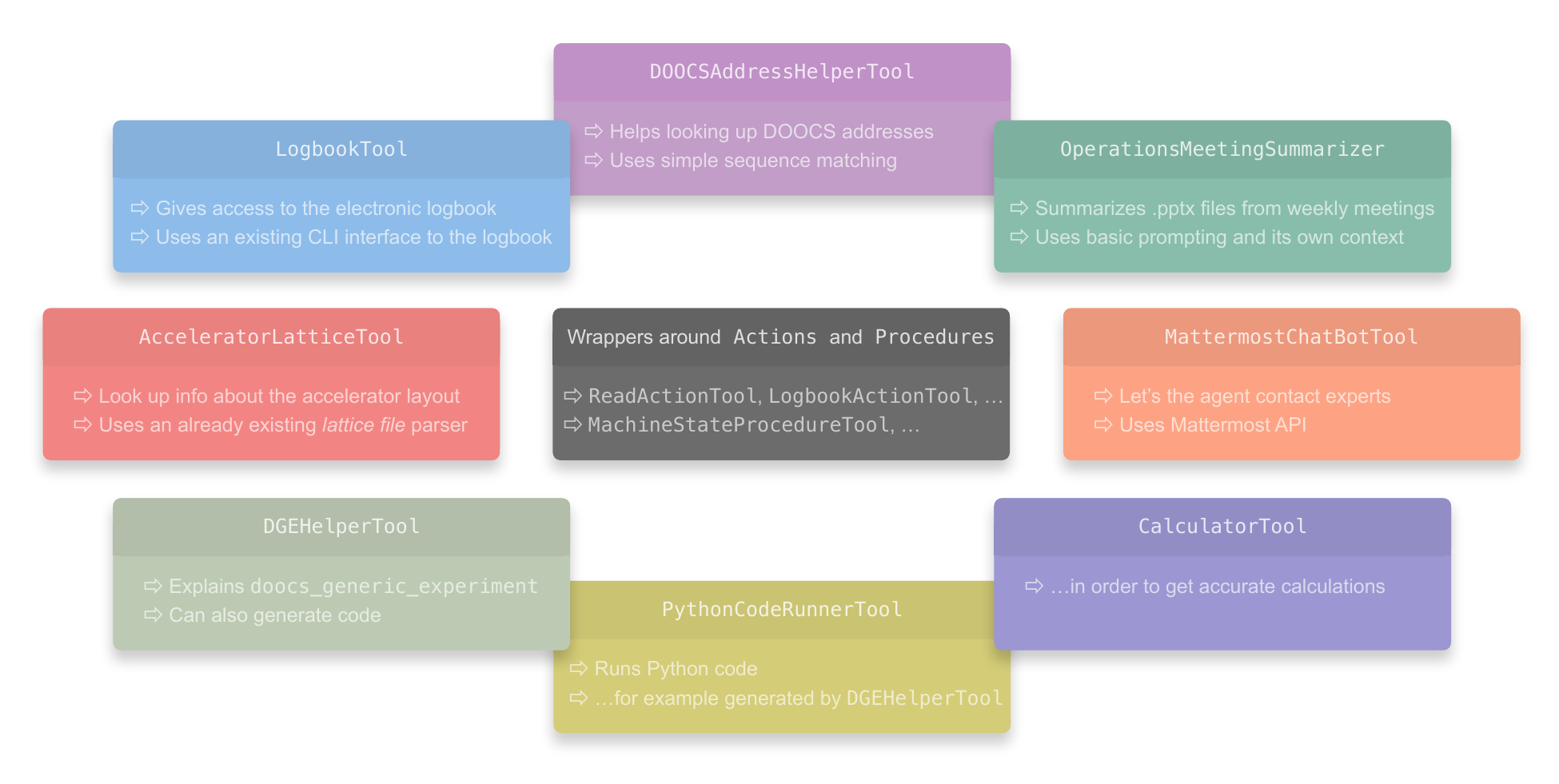}
	\caption{A selection of tools to be used by GAIA.}
	\label{fig:tools_selection}
\end{figure}

If the task is to devise a certain experimental procedure at a particle accelerator, at least two tools/experts may be needed. The first expert may provide knowledge about the layout of the beam line, as well as how to utilize the specific elements for a given task. The second expert may on the other hand be proficient in producing \texttt{doocs\_generic\_experiment} code and how to interface with the accelerator control system. The agent is able to combine the output of the two tools, yielding a suitable Python program from a prompt, such as "I want to operate the accelerator at maximum energy gain". Figure~\ref{fig:tools_selection} shows a selection of tools used in the current GAIA implementation. In the following section a number of example prompts to GAIA are shown. Note that the chain of thought, which is shown in the examples, is usually not visible to the user.

\section{Examples}
\begin{figure}[!htb]
	\centering
	\includegraphics[width=0.83\columnwidth]{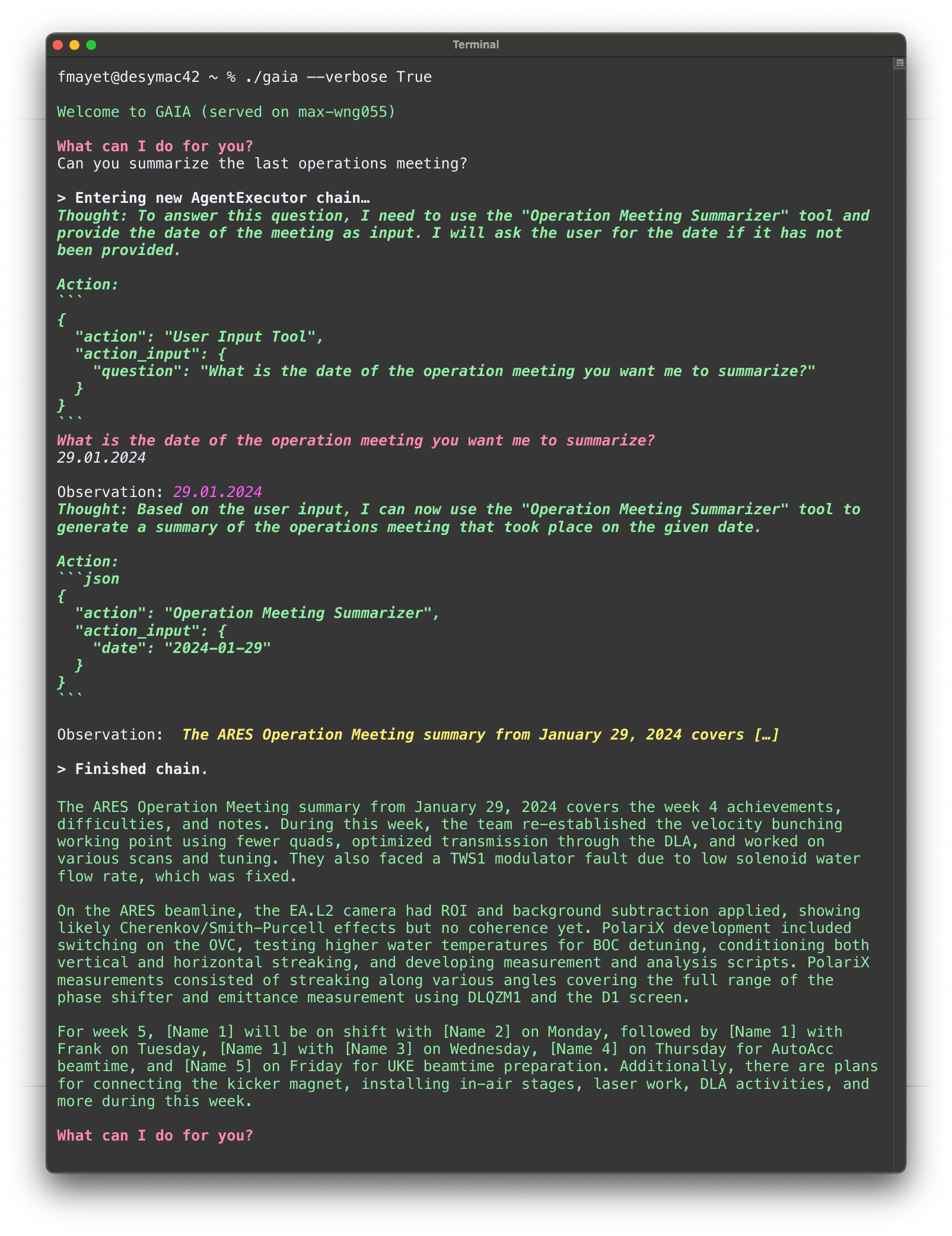}
	\caption{Example: "Can you summarize the last operations meeting?". Names are redacted.}
	\label{fig:fig1}
\end{figure}

\begin{figure}[!htb]
	\centering
	\includegraphics[width=0.83\columnwidth]{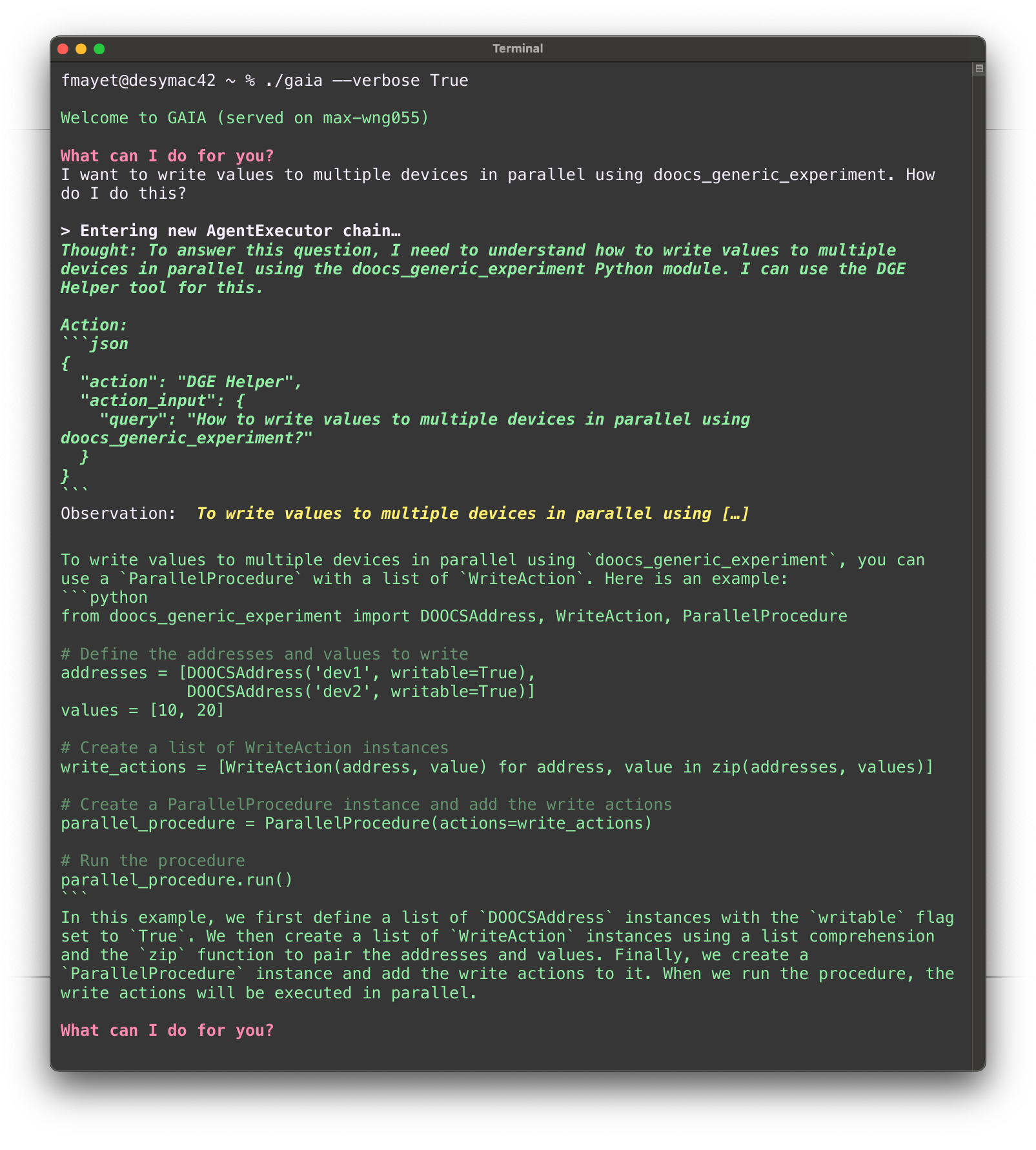}
	\caption{Example: "I want to write values to multiple devices in parallel using \texttt{doocs\_generic\_experiment}. How do I do this?".}
	\label{fig:fig2}
\end{figure}

\begin{figure}[!htb]
	\centering
	\includegraphics[width=0.83\columnwidth]{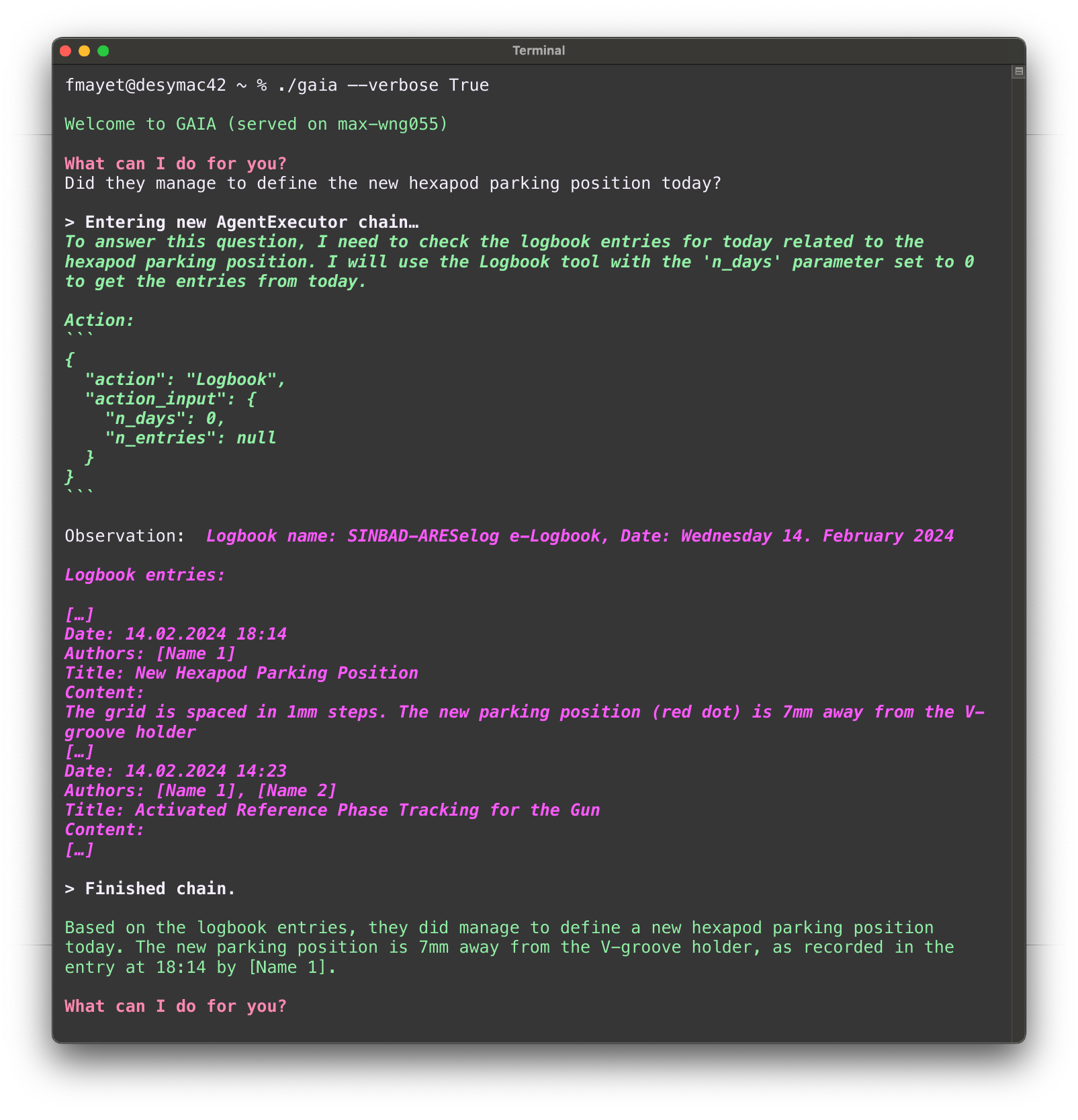}
	\caption{Example: "Did they manage to define the new hexapod parking position today?". Names are redacted.}
	\label{fig:fig3}
\end{figure}

\begin{figure}[!htb]
	\centering
	\includegraphics[width=0.46\columnwidth]{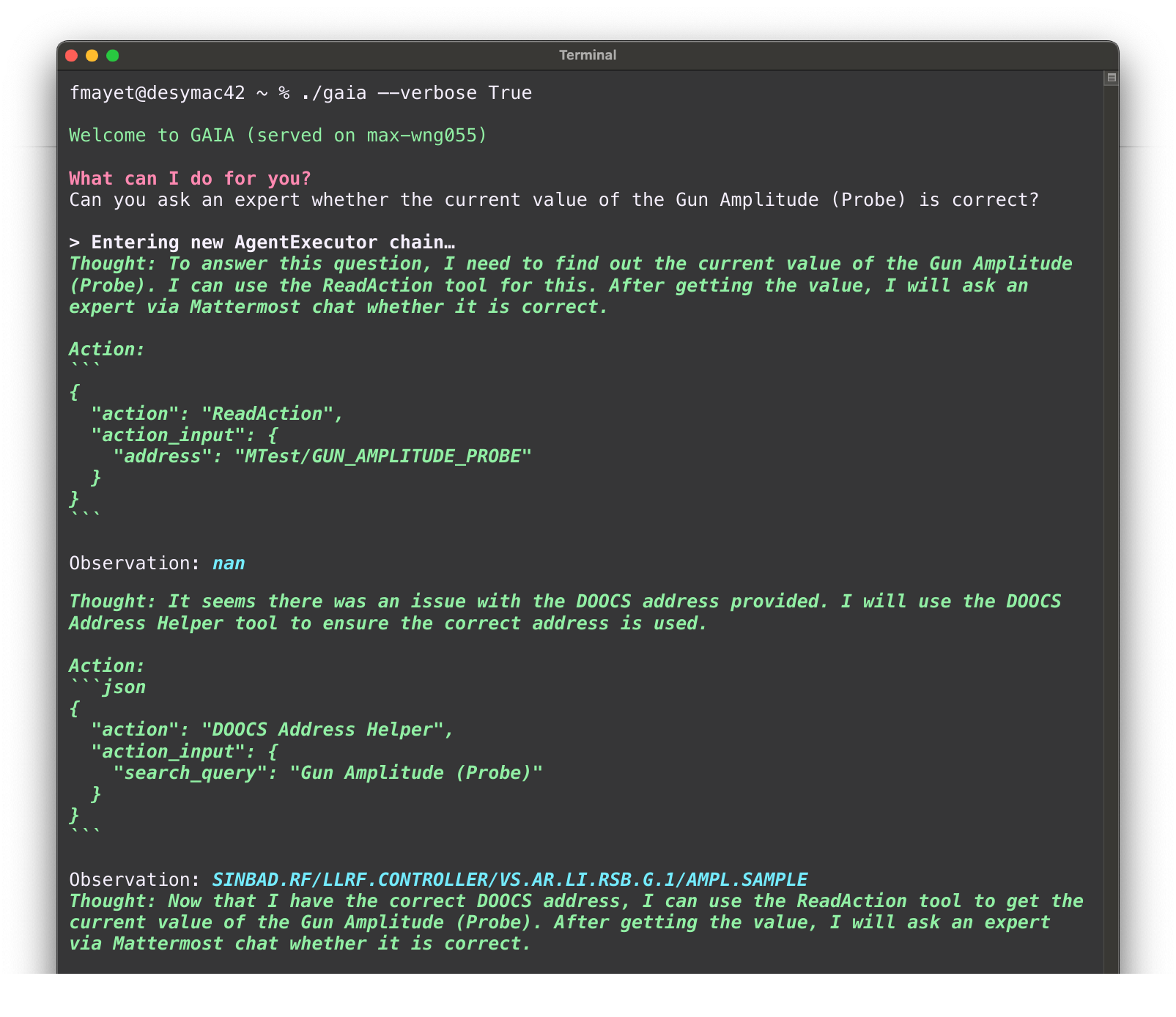}
	\includegraphics[width=0.46\columnwidth]{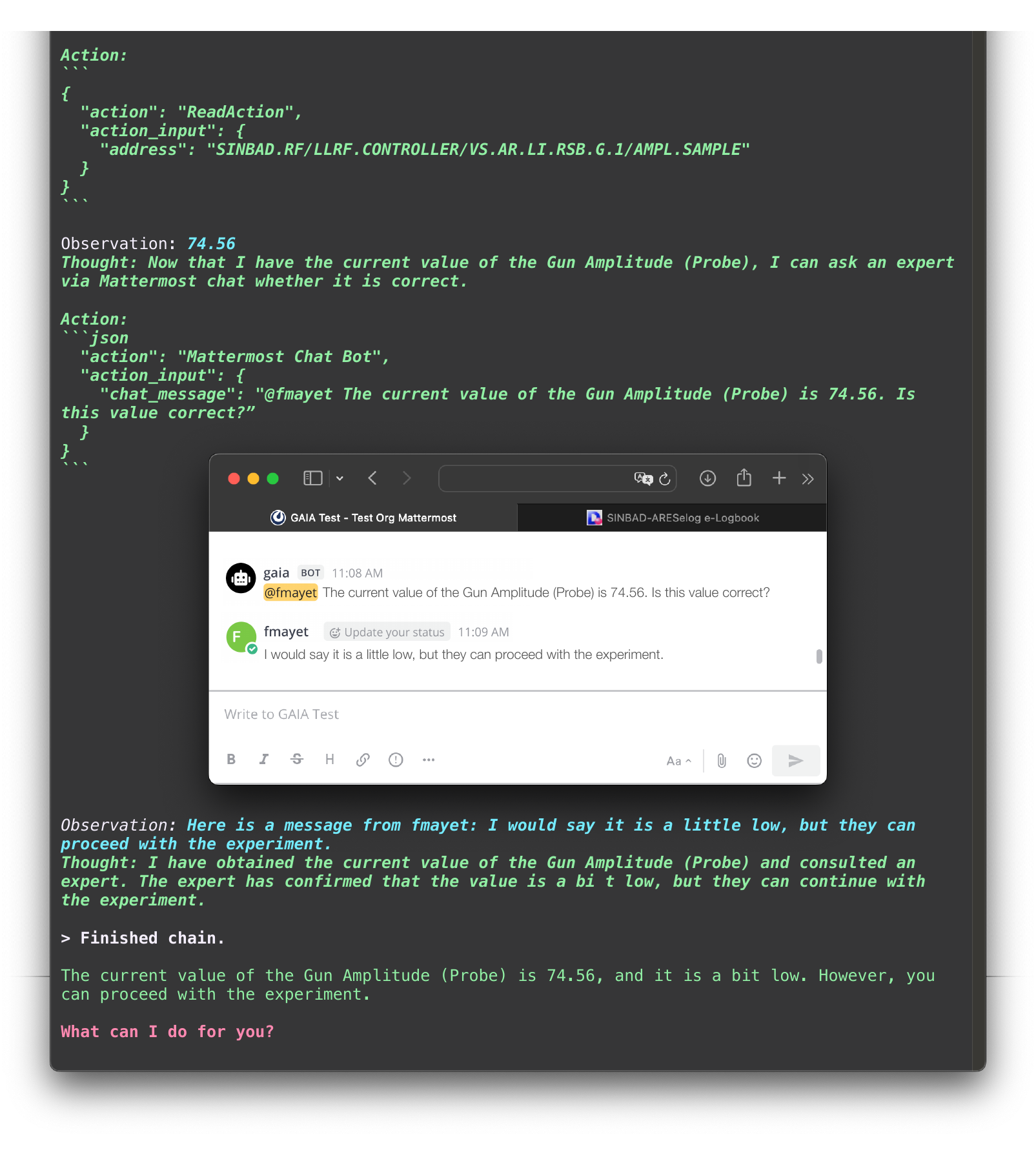}
	\caption{Example: "Can you ask an expert whether the current value of the Gun Amplitude (Probe) is correct?".}
	\label{fig:fig4}
\end{figure}

\begin{figure}[!htb]
	\centering
	\includegraphics[width=0.83\columnwidth]{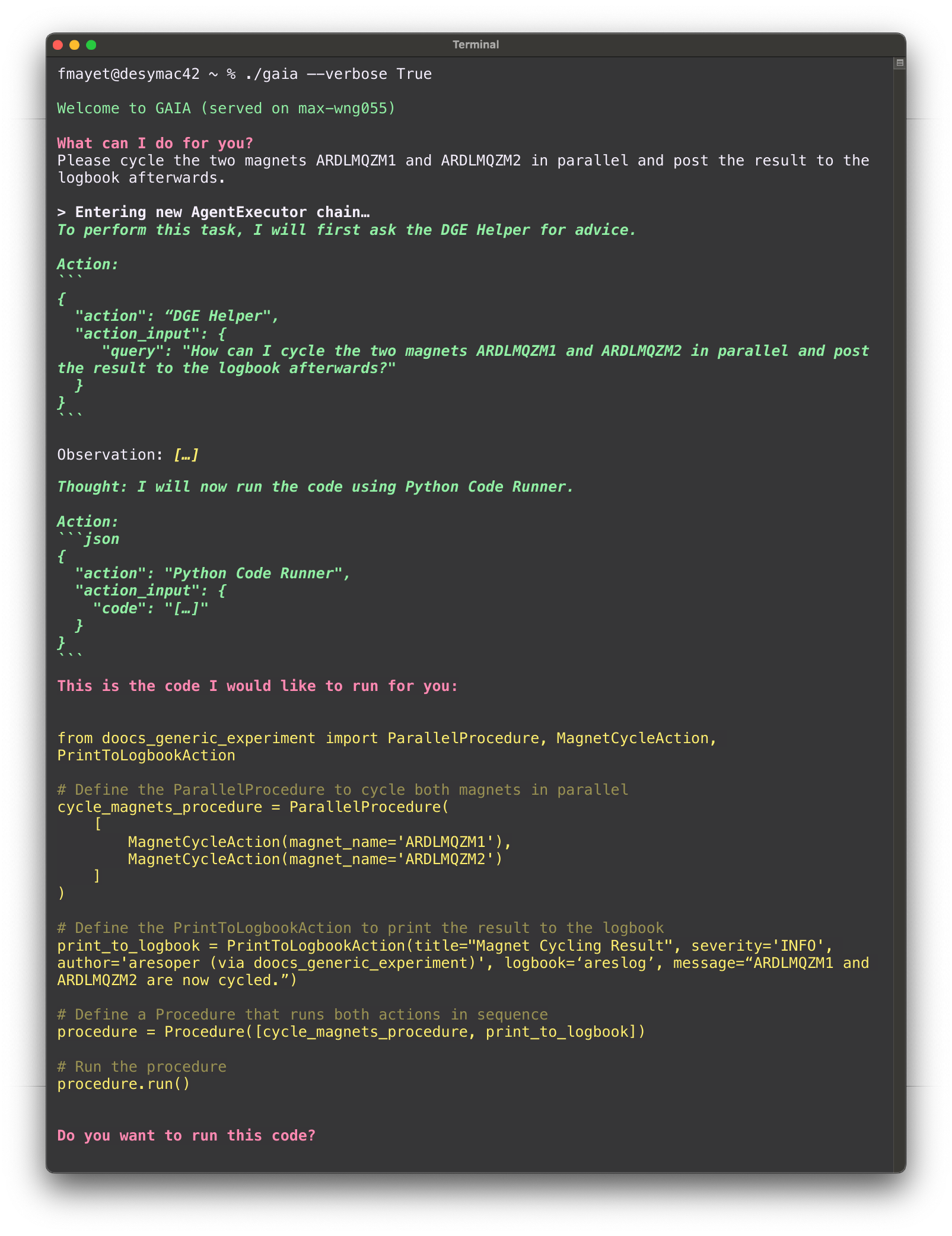}
	\caption{Example: "Please cycle the two magnets ARDLMQZM1 and ARDLMQZM2 in parallel and post the result to the logbook afterwards.".}
	\label{fig:fig5}
\end{figure}

In this section example prompts to GAIA are presented. The results are shown in Figs~\ref{fig:fig1} through \ref{fig:fig5}. Note that some parts of the output are truncated, as indicated by '[...]'. The chain of thought, shown in the examples, is usually not visible to the user. Also, names are redacted if needed.

The example cases show the versatility of the system, which includes access to meeting notes (Fig.~\ref{fig:fig1}), guidance on how to use \texttt{doocs\_generic\_experiment} (Fig.~\ref{fig:fig2}), information retrieval from the electronic logbook (Fig.~\ref{fig:fig3}), machine data access using \texttt{doocs\_generic\_experiment} and subsequently contacting human experts via Mattermost chat (Fig.~\ref{fig:fig4}), and running complex tasks using an ad hoc generated \texttt{doocs\_generic\_experiment} script (Fig.~\ref{fig:fig5}).

\section{Conclusions}
This work shows how a combination of an ReAct empowered LLM and a number of high-level expert tools can assist operators in the context of accelerator operation. The examples shown above reveal a number of positive features of the current implementation. The multi-expert nature of the ReAct concept means that one agent can be used for many use cases, which makes it easier to use for operators. At the same time, the use of tools enables save and noise-free interactions with the accelerator control system. This is crucial for machine safety. It is important to note that not all tools need to rely on advanced LLM-based systems (see Fig.~\ref{fig:tools_selection}).

As the system is work in progress, there is also room for improvement. Depending on the available compute resources the reasoning process can be lengthy, especially if many tools are used. During test runs it was furthermore observed that sometimes intermediate thoughts within the chain of thought are actually more useful than the final answer. Also, sometimes only slight changes to the initial prompt influenced the outcome substantially.

In addition to addressing and thoroughly analyzing the aforementioned issues, future work will include adding multi-modal models to better understand e.g. media rich logbook entries. Furthermore, metrics will be devised to properly quantify the quality of the answers.

\section*{Acknowledgments}
The author is grateful for getting access to the ARES R\&D accelerator at DESY, as well as being able to leverage the compute resources provided by the DESY Maxwell team. The author acknowledges support from DESY (Hamburg, Germany), a member of the Helmholtz Association HGF.

\bibliographystyle{unsrtnat}
\bibliography{gaia_paper}  

\end{document}